\documentclass[journal]{IEEEtranTIE}
\usepackage{graphicx}
\usepackage{cite}
\usepackage{picinpar}
\usepackage{amsmath}
\usepackage{url}
\usepackage{flushend}
\usepackage{colortbl}
\usepackage{soul}
\usepackage{multirow}
\usepackage{pifont}
\usepackage{color}
\usepackage[table,xcdraw]{xcolor}
\usepackage{xcolor}
\usepackage{alltt}
\usepackage[hidelinks]{hyperref}
\usepackage{enumerate}
\usepackage{siunitx}
\usepackage{breakurl}
\usepackage{epstopdf}
\usepackage{pbox}
\usepackage{amsmath,amsfonts}
\usepackage{algorithmic}
\usepackage{algorithm}
\usepackage{array}
\usepackage[caption=false,font=normalsize,labelfont=sf,textfont=sf]{subfig}
\usepackage{textcomp}
\usepackage{stfloats}
\usepackage{url}
\usepackage{verbatim}
\usepackage{graphicx}
\usepackage{cite}
\usepackage{breqn}
\usepackage{makecell}
\usepackage{tabularx}
\usepackage{adjustbox}
\usepackage{amsmath,amsfonts}
\usepackage{algorithmic}
\usepackage{algorithm}
\usepackage{array}
\usepackage[caption=false,font=normalsize,labelfont=sf,textfont=sf]{subfig}
\usepackage{textcomp}
\usepackage{stfloats}
\usepackage{url}
\usepackage{verbatim}
\usepackage{graphicx}
\usepackage{cite}
\usepackage{graphicx}
\usepackage{cite}
\usepackage{picinpar}
\usepackage{amsmath}
\usepackage{url}
\usepackage{flushend}
\usepackage{colortbl}
\usepackage{soul}
\usepackage{multirow}
\usepackage{pifont}
\usepackage{color}
\usepackage[table,xcdraw]{xcolor}
\usepackage{xcolor}
\usepackage{alltt}
\usepackage[hidelinks]{hyperref}
\usepackage{enumerate}
\usepackage{siunitx}
\usepackage{breakurl}
\usepackage{epstopdf}
\usepackage{pbox}
\usepackage{amsmath,amsfonts}
\usepackage{algorithmic}
\usepackage{algorithm}
\usepackage{array}
\usepackage[caption=false,font=normalsize,labelfont=sf,textfont=sf]{subfig}
\usepackage{textcomp}
\usepackage{stfloats}
\usepackage{url}
\usepackage{verbatim}
\usepackage{graphicx}
\usepackage{cite}
\usepackage{breqn}
\usepackage{makecell}
\usepackage[utf8]{inputenc} 
\usepackage{tabularx}
\begin{document}
\title{A Miniature High-Resolution Tension Sensor Based on a Photo-Reflector for Robotic Hand and Grippers}

\author{                                                                      
	\vskip 1em
	
	Hyun-Bin Kim, and Kyung-Soo Kim,~\IEEEmembership{Member,~IEEE,}

	\thanks{
	Manuscript prepared in March 2025; this work was developed by Mechatronics, Systems and Control (MSC) laboratory, Korea Advanced Institute of Science and Technology (KAIST), Daehak-Ro 291, Daejeon, South Korea (email: youfree22@kaist.ac.kr; kyungsookim@kaist.ac.kr).(Corresponding author: Kyung-Soo Kim). 
	}
}

\maketitle
	
\begin{abstract}
This paper presents a miniature tension sensor using a photo-reflector, designed for compact tendon-driven grippers and robotic hands. The proposed sensor has a small form factor of 13~mm × 7~mm × 6.5~mm and is capable of measuring tensile forces up to 200~N. A symmetric elastomer structure incorporating fillets and flexure hinges is designed based on Timoshenko beam theory and verified via FEM analysis, enabling improved sensitivity and mechanical durability while minimizing torsional deformation. The sensor utilizes a compact photo-reflector (VCNT2020) to measure displacement in the near-field region, eliminating the need for light-absorbing materials or geometric modifications required in photo-interrupter-based designs. A 16-bit analog-to-digital converter (ADC) and CAN-FD (Flexible Data-rate) communication enable efficient signal acquisition with up to 5~kHz sampling rate. Calibration experiments demonstrate a resolution of 9.9~mN (corresponding to over 14-bit accuracy) and a root mean square error (RMSE) of 0.455~N. Force control experiments using a twisted string actuator and PI control yield RMSEs as low as 0.073~N. Compared to previous research using photo-interrupter, the proposed method achieves more than tenfold improvement in resolution while also reducing nonlinearity and hysteresis. The design is mechanically simple, lightweight, easy to assemble, and suitable for integration into robotic and prosthetic systems requiring high-resolution force feedback.

\end{abstract}


\markboth{arXiv}%
{}

\definecolor{limegreen}{rgb}{0.2, 0.8, 0.2}
\definecolor{forestgreen}{rgb}{0.13, 0.55, 0.13}
\definecolor{greenhtml}{rgb}{0.0, 0.5, 0.0}

\section{Introduction}
\label{sec:introduction}
\IEEEPARstart{I}{n} recent years, robotic arms and robotic fingers have been rapidly developed and commercialized~\cite{zhang2025biomimetic,shahriari2024path}. Research in the field of prosthetics has also been actively conducted, leading to the introduction of various products. For robots and devices that interact closely with humans-such as collaborative robots-force/torque sensors are commonly integrated~\cite{ding2021situ,kim2021novel}. This is because performing position control without force sensing can potentially cause harm to users. Therefore, force-related sensors are of critical importance in such applications.

As humanoid robots continue to evolve~\cite{radosavovic2024real}, finger-shaped grippers have advanced accordingly and are being employed in a wide range of robotic systems~\cite{voigt2025learning}. These grippers must grasp objects without causing damage, which necessitates precise force sensing. Likewise, in prosthetic hands, accurate tension sensing is essential for safely and intuitively replicating human grasp behavior. Accordingly, force/torque sensors or tactile sensors~\cite{lu2022gtac,liu2024material} are often integrated into robotic grippers and myoelectric prostheses to enable both precision and safety.

Conventional grippers use various actuation mechanisms, such as gear-driven systems~\cite{zuo2021grasping} and tendon-based systems~\cite{zhao2023adaptive,cho2022msc}. In this paper, we propose a miniature tension sensor suitable for tendon-driven mechanisms or twisted string actuators (TSA). 

While strain-gauge-based methods are commonly used for tension measurement~\cite{da2002strain,qu20203d,wei2015overview}, they require direct attachment of the gauge to the elastomer and the use of an external amplifier, making them less suitable for compact applications as grippers and robotic hands. In addition, most conventional load cells also rely on strain-gauge technology, typically requiring external amplifiers and supply voltages exceeding 12~V. 

In contrast, the proposed sensor utilizes a photo-reflector to measure displacement. Unlike previous studies that employed photo-interrupters-which rely on light absorption-the photo-reflector measures displacement based on light reflection. As a result, the proposed method does not require additional calibration processes such as applying light-absorbing materials or modifying the screen geometry. Moreover, photo-interrupter-based methods typically have a narrow measurement range and require precise calibration, which is difficult to perform reliably.

Related works include the application of photo-reflectors to six-axis force/torque sensors~\cite{kim2024compact, kim2025parameter, kim2025temperature, kim2023compact}, where methods for temperature compensation have been studied. There are also approaches using photo-reflectors in conjunction with 3D printing~\cite{noh2016multi} and finite element analysis to derive optimized sensor designs.

The target application of this study is a gripper or robotic hand driven by a TSA~\cite{shin2012robot}, which is a compact actuation mechanism capable of generating forces up to 200~N. Therefore, the allowable force target for the proposed sensor was set to 200~N~\cite{cho2022msc}.

This paper focuses on the design and evaluation of a compact tension sensor based on a photo-reflector. The primary emphasis is placed on the characteristics and performance of the sensor itself, with an eye toward future integration into wearable prosthetic devices and high-performance robotic grippers.

\subsection{Main Contributions}

The proposed method presents several key advantages over previous approaches:

\begin{itemize}
    \item \textbf{High Resolution:} The proposed sensor achieves a resolution of approximately 0.01~N under a 200~N load. This is more than ten times greater than the 0.1~N resolution reported in previous studies. Such high resolution is particularly critical in applications like prosthetic hands, where subtle force feedback is needed for natural object manipulation.

    \item \textbf{Large Displacement with Minimal Angular Variation:} Unlike conventional C-shaped spring designs, which deform not only in displacement but also in angle-thereby affecting the accuracy of displacement measurement-the proposed spring structure is laterally symmetric. This symmetry ensures that the deformation occurs primarily in translation, minimizing angular distortion and improving measurement consistency, which is essential for stable control in robotic grippers.
    \item \textbf{Effective Closed-loop Force Control:} Experimental results demonstrate the feasibility of high-precision force control using a TSA, achieving an root mean square error (RMSE) of 0.073 N with a simple PI controller.
\end{itemize}

\section{Design and Operating Principle of the Proposed Tension Sensor}
\begin{figure}[!htb]
    \centering
    \includegraphics[width=0.9\linewidth]{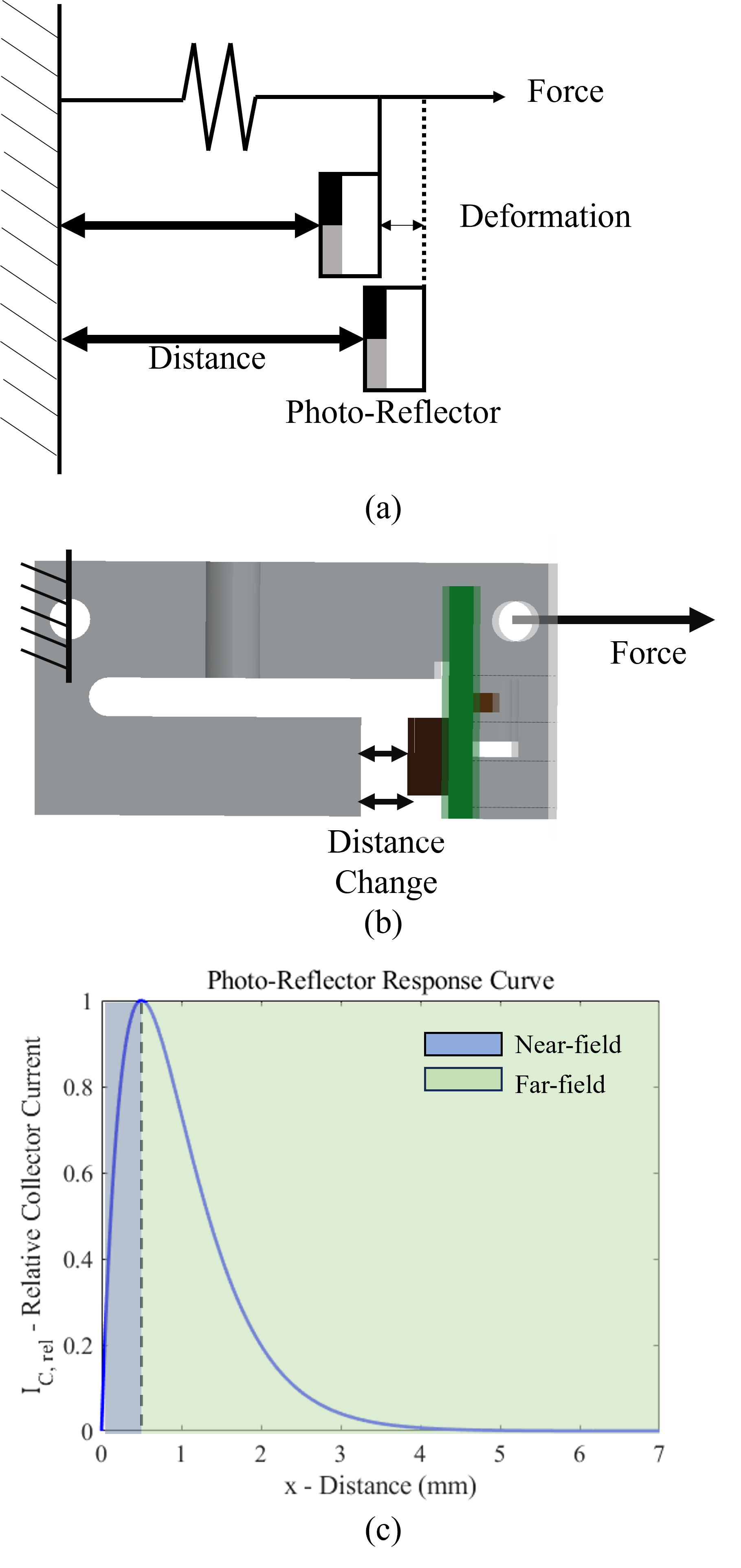}
    \caption{
    Principle of the proposed sensor: (a) Conceptual illustration of the sensor's deformation when external tensile force is applied, showing how displacement is generated in the elastomer structure; (b) Graphical representation of the proposed sensor design, including the relative placement of the photo-reflector and reflective surface; (c) Characteristic response curve of the photo-reflector showing the variation in output current as a function of the distance between the emitter and the reflective surface, which forms the basis of the displacement-to-voltage conversion in the sensing mechanism.}
    \label{principle}
\end{figure}
The working principle of the proposed sensor is straightforward. A photo-reflector outputs an analog voltage that varies with the distance to a reflective surface. By exploiting this property, the sensor measures displacement resulting from elastic deformation when a force is applied, as illustrated in Fig.~\ref{principle} (a). Specifically, as shown in Fig.~\ref{principle} (b), the left hole is fixed with a pin, while a string is attached to the right hole to apply tension. When a tensile force is exerted, the elastic structure stretches, leading to a change in the distance detected by the photo-reflector. This enables the measurement of tension as a corresponding analog voltage output.

As depicted in Fig.~\ref{principle} (c), the photo-reflector exhibits two characteristic regions: a \textit{near-field} region where the output voltage increases sharply with decreasing distance, and a \textit{far-field} region where the voltage gradually decreases after reaching a peak. Unlike previous studies on six-axis force/torque sensors, which utilized the far-field region, this study focuses on utilizing the near-field region, which lies approximately in the range of 0.2--0.5~mm.

\begin{figure}[!htb]
    \centering
    \includegraphics[width=0.8\linewidth]{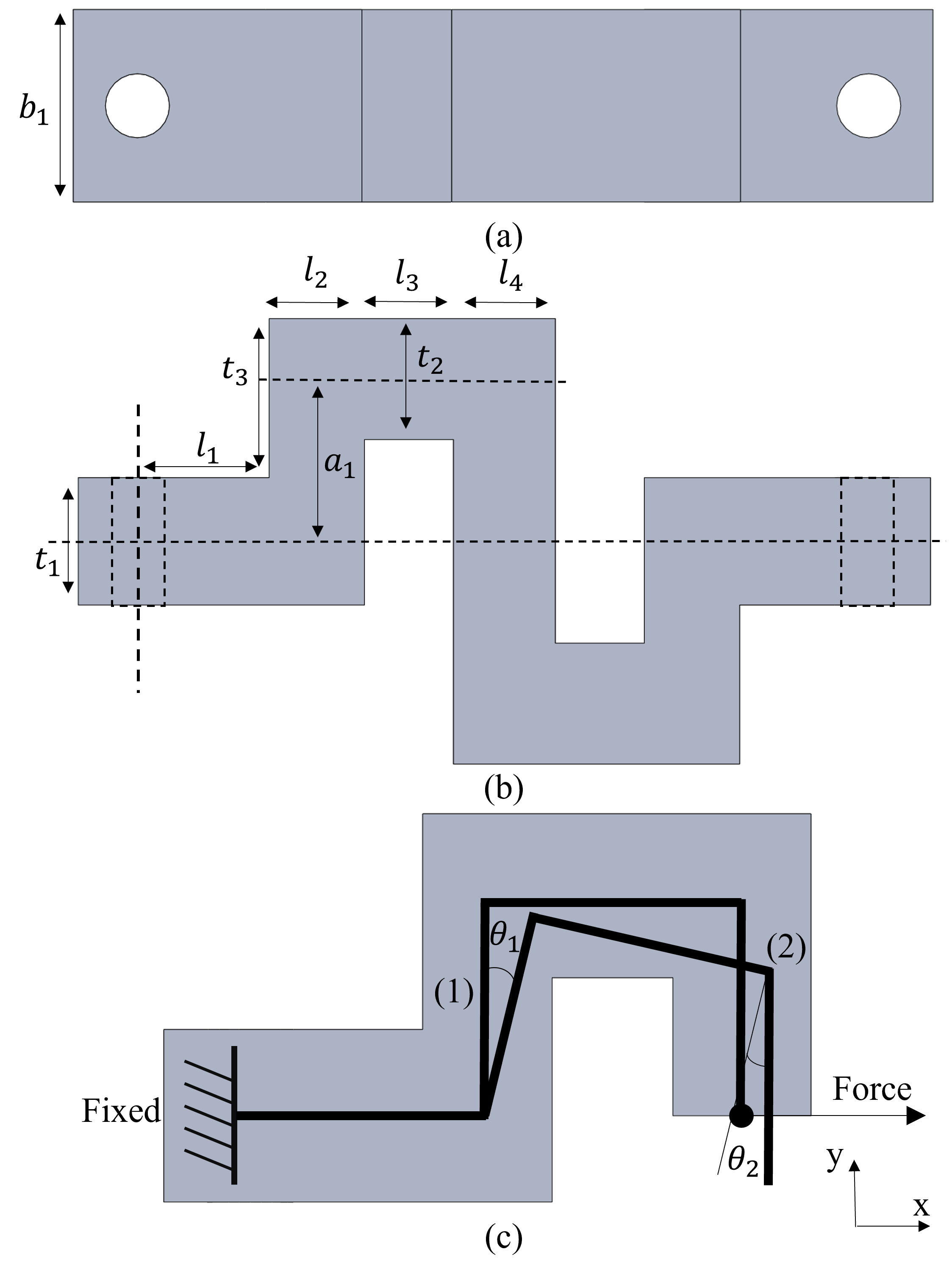}
    \caption{
    Structure of the proposed sensor’s elastomer: (a) Top view of the elastomer, primarily included to present the thickness distribution and planform geometry relevant to mechanical design considerations; (b) Front view of the elastomer, illustrating the overall geometry and key structural dimensions that were used in the design and fabrication process; (c) Analytical model representing half of the elastomer geometry, used for theoretical deformation and stress analysis based on Timoshenko beam theory.}
    \label{structure}
\end{figure}
Fig.~\ref{structure} illustrates the fundamental design of the elastomer used in the proposed sensor. Fig.~\ref{structure} (a) shows the \textit{top view}, where the thickness of the elastomer, denoted as $b_1$, can be observed. To maintain symmetry and ensure that the vertical position along the $y$-axis remains unchanged when tension is applied from both ends, the height $b_1$ is set to be equal on both sides.

Fig.~\ref{structure} (b) presents the \textit{front view}, in which various dimensional parameters of the elastomer can be seen, including the lengths of the horizontal beams $l_1$, $l_2$, $l_3$, $l_4$, thicknesses $t_1$, $t_2$, and the vertical spacing $a_1$.

Fig.~\ref{structure} (c) depicts a simplified model for structural analysis. Due to the symmetry of the design, the structure is halved for analysis purposes, with the left side fixed and the right side subjected to an external tensile force. To mathematically model this configuration, the Timoshenko beam theory can be applied. Owing to the geometric simplicity of the structure, the analytical formulation is straightforward.

The material selected for the elastomer is AL7075-T6, which has low silicon content, is lightweight, and exhibits minimal hysteresis. The elastic modulus $E$ is set to 7.17~GPa, and the shear modulus $G$ is set to 26.9~GPa. A shear coefficient $k_s$ of $5/6$ is used in the analysis.

\begin{equation}
    \delta_x = 2(\delta_1+\delta_2+\delta_3+\delta_4+\delta_5+\delta_6)
\end{equation}
Here, $\delta_1$, $\delta_2$, $\delta_3$, and $\delta_4$ represent the displacements caused by tensile force in the segments $l_1$, $l_2$, $l_3$, and $l_4$, respectively. These displacements (m) can be expressed using the classical axial deformation formula:
\[
\delta_{axial} = \frac{FL}{EA}
\]
where $F$ is the applied force (N), $L$ is the length of the beam segment (m), $E$ is the elastic modulus (N/$\text{m}^2$), and $A$ is the cross-sectional area ($\text{m}^2$).

The total axial displacement (m) can be expressed as:
\begin{equation}
    \delta_1+\delta_2+\delta_3+\delta_4 = \frac{F l_1}{EA_1}+\frac{F l_2}{EA_2}+\frac{F l_3}{EA_3}+\frac{F l_4}{EA_4}
\end{equation}
where the cross-sectional areas are defined as follows: $A_1 = b_1 t_1$, $A_2 = b_1(t_1 + t_3)$, $A_3 = b_1 t_2$, and $A_4 = b_1(\frac{t_1}{2} + t_3)$.

In addition, the displacements due to shear deformation, denoted by $\delta_5$ and $\delta_6$, can be calculated based on Timoshenko beam theory~\cite{pu2021modeling}, which accounts for angular deflection due to shear forces. Here, we assume that the angular displacements $\theta_1$ and $\theta_2$ are small, allowing the approximation $\cos(\theta_1) \approx 1$ and $\cos(\theta_2) \approx 1$.

Under this assumption, the displacements can be expressed as:
\begin{equation}
    \delta_5 = a_1 \theta_1, \quad \delta_6 = a_1(\theta_2 - \theta_1)
\end{equation}

Accordingly, the expressions for $\delta_5$ and $\delta_6$ can be reformulated as:
\begin{equation}
    \delta_5 = F a_1^2 \left( \frac{a_1}{EI_1} + \frac{1}{k_s G A_5} \right)
\end{equation}
\begin{equation}
    \delta_6 = F a_2 \left[ a_2 \left( \frac{a_2}{EI_2} + \frac{1}{k_s G A_6} \right) - a_1 \left( \frac{a_1}{EI_1} + \frac{1}{k_s G A_5} \right) \right]
\end{equation}

By substituting the values from Table~\ref{dimension} into the equation and applying a force of 200~N, the resulting displacement $\delta_x/2$ is approximately 0.04753~mm. This result closely matches the displacement obtained from the Finite Element Method (FEM) analysis in SOLIDWORKS, which is 0.04984~mm. However, due to the difficulty of machining sharp corners, fillets were introduced, and the structure was designed to withstand a load of 200~N. 

\begin{table}[!htb]
    \centering
    \caption{Dimension of Elastomer}
    \begin{tabular}{cccccccccc}
    \hline\hline
        Dimension & $l_1$&$l_2$&$l_3$&$l_4$&$t_1$&$t_2$&$t_3$&$a_1$&$b_1$ \\\hline
        Value(mm) &2 &1.5&1.4&1.6&2&1.9&2.5&2.55&3\\\hline
    \end{tabular}
    
    \label{dimension}
\end{table}

\begin{figure}[!htb]
    \centering
    \includegraphics[width=1\linewidth]{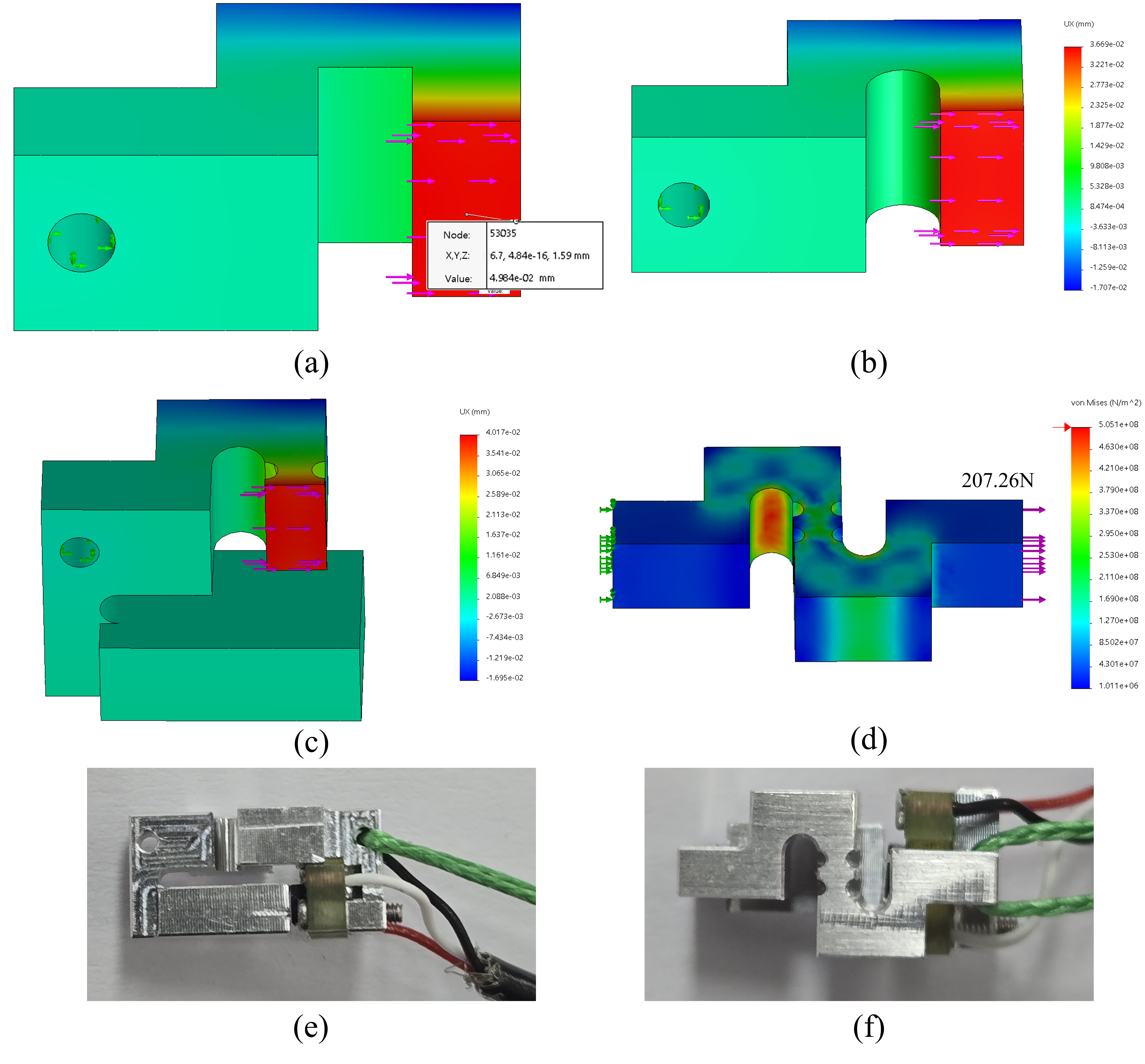}
    \caption{FEM simulation results and fabricated proposed sensor: (a) FEM result of the initial design without a fillet, showing higher localized stress concentrations; (b) Simulation result after introducing a fillet, demonstrating reduced stress and displacement; (c) Enhanced design incorporating a central flexure hinge along with the fillet structure to improve sensitivity while maintaining structural integrity; (d) FEM simulation showing the maximum bearable force of the sensor, confirmed to be 207.26~N; (e) Front view of the fabricated sensor, illustrating the integration of the elastomer structure with the PCB assembly; (f) Top view of the fabricated sensor, highlighting the fillet and flexure hinge geometry implemented in the final design.}
    \label{fem}
\end{figure}

Fig.~\ref{fem} presents the FEM simulation results along with photographs of the fabricated sensor. In Fig.~\ref{fem} (a), the displacement is approximately 0.04984~mm, which closely matches the theoretical value calculated using Timoshenko beam theory. When a fillet is introduced, the displacement decreases to 0.03669~mm as shown in Fig.~\ref{fem} (b). By incorporating two flexure hinges, as illustrated in Fig.~\ref{fem} (c), the sensitivity can be enhanced, resulting in a displacement of 0.04017~mm, representing an increase of approximately 10\%. Fig.~\ref{fem} (d) demonstrates that the sensor can withstand a maximum applied force of 207.26~N. Fig.~\ref{fem} (e) shows the side view of the fabricated sensor, where the elastomer is integrated with the printed circuit board (PCB). Similarly, Fig.~\ref{fem} (f) shows the top view of the sensor, highlighting the presence of both the fillet and the flexure hinge.

\begin{table}[!htb]
    \centering
    \caption{Physical Properties of Proposed Sensor}
    \begin{tabular}{cc}
    \hline\hline
    Physical Properties & Value\\ \hline
      Dimension   &  13~mm$\times$7~mm$\times$6.5 ~mm\\ 
    Weight         & 0.8~g\\
    Material & AL7075-T6\\
   
    \hline
    \end{tabular}
    
    \label{phy}
\end{table}

The physical characteristics of the proposed sensor are summarized in Table~\ref{phy}. The sensor has a width of 7~mm, a length of 13~mm, and a height of 6.5~mm, with a total weight of approximately 0.8~g, indicating that it is extremely lightweight.

The sensor consists of two main components: an elastomer and a PCB. The PCB integrates a compact photo-reflector, the VCNT2020 from Vishay, which has a size of 2.5~mm.

\begin{figure}[!htb]
    \centering
    \includegraphics[width=1\linewidth]{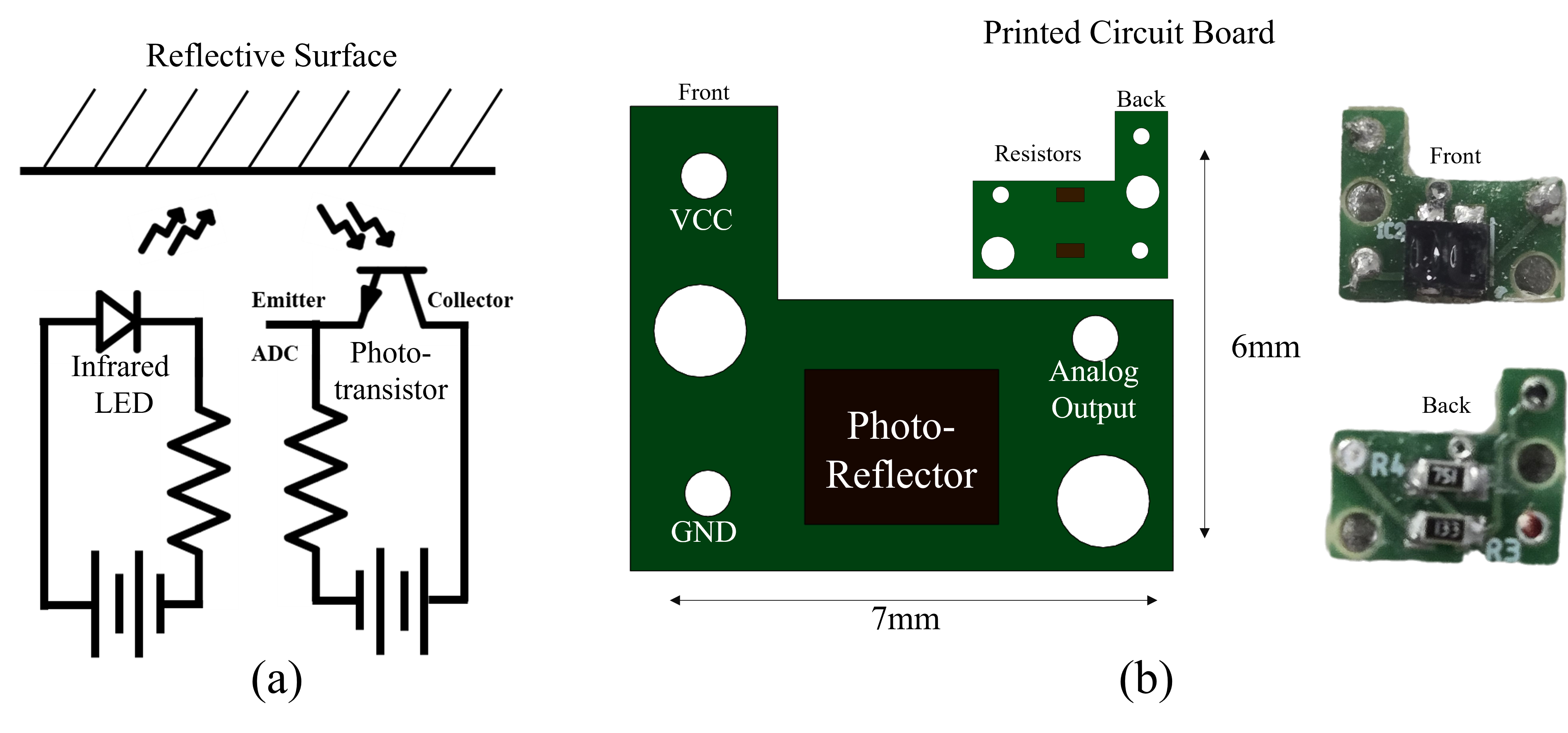}
    \caption{PCB overview: (a) circuit diagram of the photo-reflector, (b) actual PCB layout and photograph.}
    \label{PCB}
\end{figure}

Fig.~\ref{PCB} illustrates the circuit design and components of the PCB, as well as the actual soldered PCB. Fig.~\ref{PCB} (a) shows the internal structure of the photo-reflector, which consists of an infrared LED and a phototransistor. The LED emits infrared light, while the phototransistor outputs an analog signal from its emitter. This analog output is connected to the ADC input. As shown in Fig.~\ref{PCB} (b), the circuit utilizes three lines: two for power supply to the photo-reflector and one for the analog output. The design includes only two resistors, one for the LED and the other for the phototransistor. Screw holes are placed on both sides to allow mechanical fixation to the elastomer using bolts instead of adhesive. Unlike previous sensors that required adhesive bonding, the use of bolts provides a more flexible and reliable assembly method. Furthermore, while conventional designs often employed photo-interrupters-necessitating the use of light-absorbing materials or specialized geometries to minimize diffuse reflection-the use of photo-reflectors in the proposed design reduces such considerations, simplifying the sensor structure.

\section{Calibration and Force Control Experiment Using Motor}
\begin{figure}[!htb]
    \centering
    \includegraphics[width=1\linewidth]{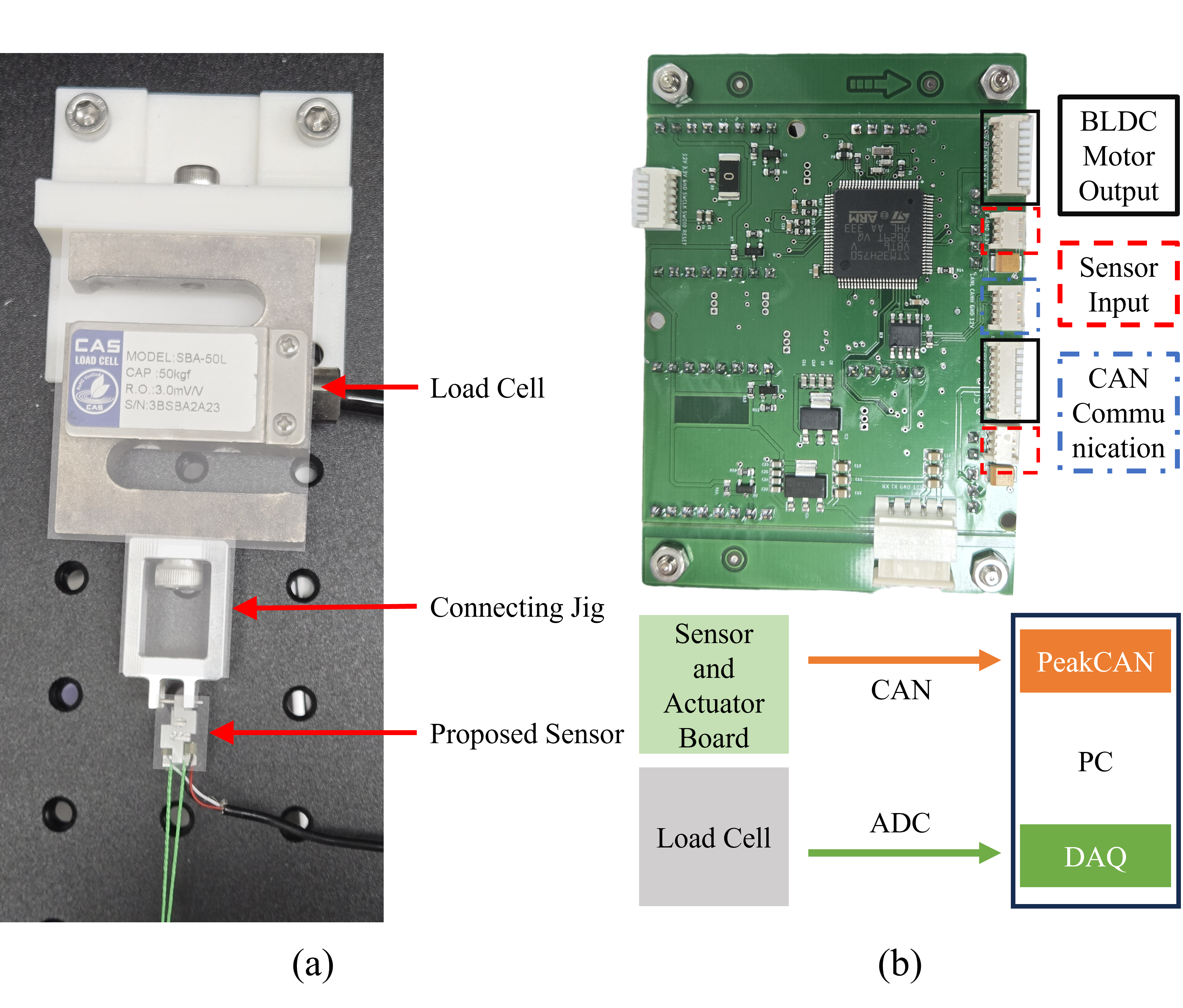}
    \caption{Calibration experiment setup: (a) direct connection of the load cell and the proposed sensor for calibration, (b) data acquisition using a PC and a PCB that interfaces the sensor and motor for calibration and force control experiments.}
    \label{calexpins}
\end{figure}

Next, calibration and performance evaluation experiments were conducted for the sensor. For calibration, a load cell was used. Specifically, a CAS SBA-50L load cell capable of measuring up to 50~kgf was employed. The load cell provides an output of 3.0~mV/V and was connected to a 12~V amplifier. Data acquisition (DAQ) was performed using the NI-PCIE 6363 model, and the load cell output was collected at a sampling rate of 1~kHz using Simulink Desktop Real-Time.

The proposed sensor was interfaced with an STM32H750 microcontroller unit (MCU), which utilized a 16-bit ADC for signal measurement. The acquired data was transmitted to a PC via CAN communication. On the PC side, the Peak system’s PCAN M.2 interface was used to log data at a sampling rate of 1~kHz. The proposed sensor requires only three connections to the MCU: 3.3~V power, ground, and analog output, which is sufficient to acquire signals.

Fig.~\ref{calexpins} shows the calibration setup and the PCB responsible for sensor input. In Fig.~\ref{calexpins} (a), the load cell is mounted on a granite surface plate, and a connection jig is placed between the load cell and the proposed sensor to create a serial configuration. This configuration is commonly used in previous studies.

Fig.~\ref{calexpins} (b) presents a photograph of the PCB that processes the sensor input and transmits the data to a PC via CAN communication, along with a diagram of the communication setup for the calibration system. CAN-FD (Flexible Data-rate) was used for communication, enabling transmission rates of up to 5~Mbps. This board was also designed for use in subsequent experiments involving motor-driven force application.
The power and signal lines of the proposed sensor were connected using a shielded wire with a length of approximately 40~cm.

\begin{figure*}[!htb]
    \centering
    \includegraphics[width=1\linewidth]{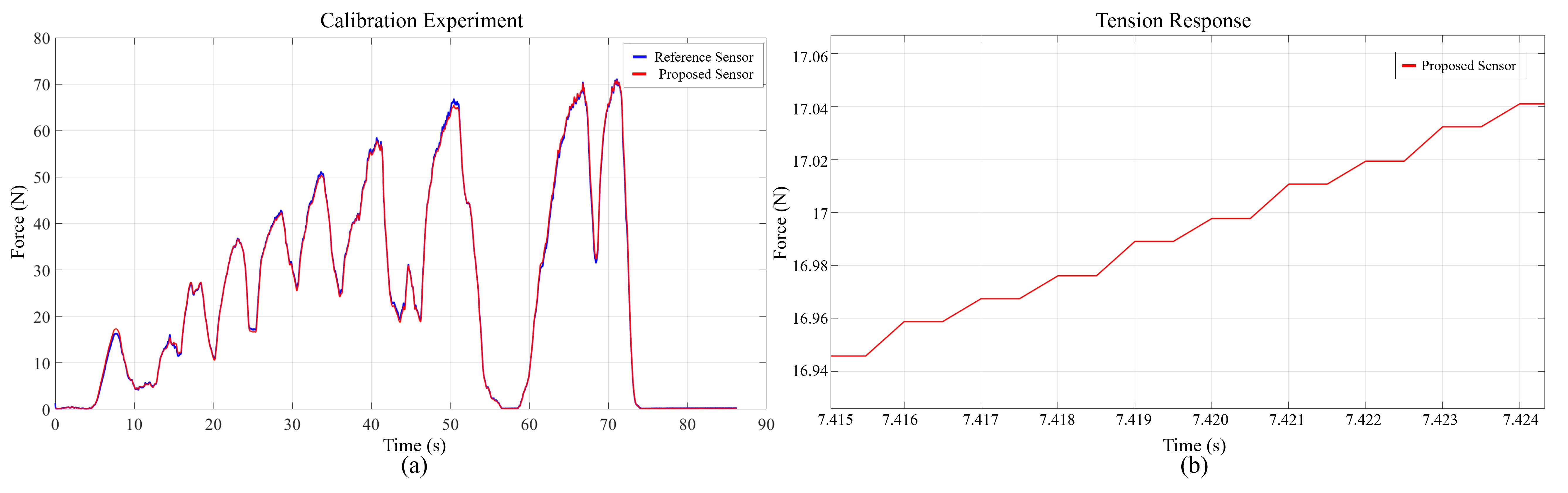}
    \caption{Calibration result graphs: (a) Graph showing the overall calibration result obtained using a third-order polynomial fitting method, demonstrating the nonlinear response of the sensor under applied tensile force; (b) Magnified view of the low-force region used to verify the sensor's resolution, showing discrete measurement steps corresponding to a resolution of approximately 9.9~mN.}
    \label{calibrationexperiment}
\end{figure*}

\begin{figure}[!htb]
    \centering
    \includegraphics[width=1\linewidth]{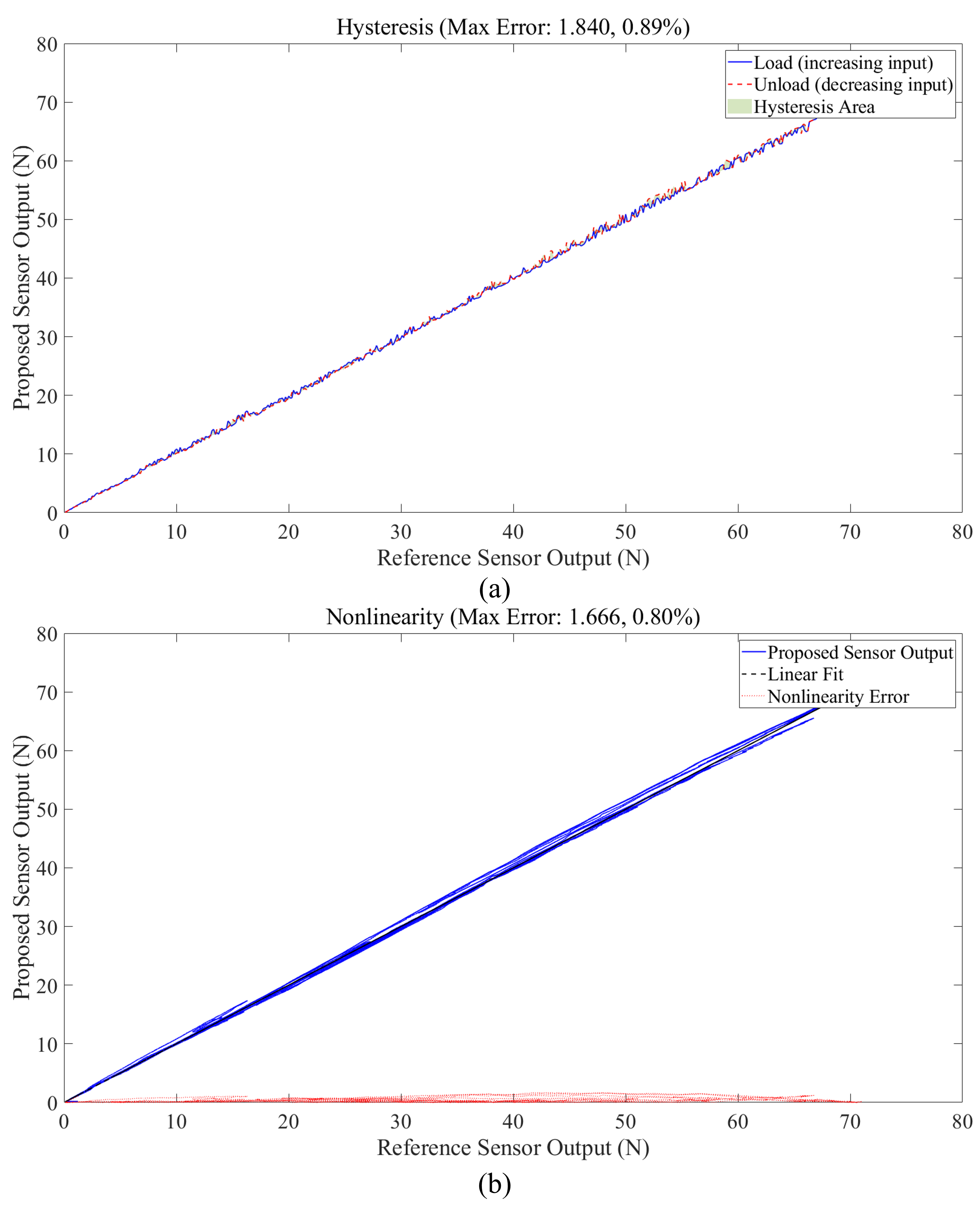}
    \caption{Hysteresis and Nonlinearity analysis: (a) Hysteresis curve illustrating the difference in sensor output between increasing and decreasing force inputs, highlighting the minimal hysteresis behavior of the proposed sensor; (b) Nonlinearity plot showing the deviation of the sensor response from the ideal linear fit across the full force range, used to evaluate the linearity characteristics of the sensor under tensile loading.}
    \label{nonlinear}
\end{figure}
\begin{figure}[!tb]
    \centering
    \includegraphics[width=1\linewidth]{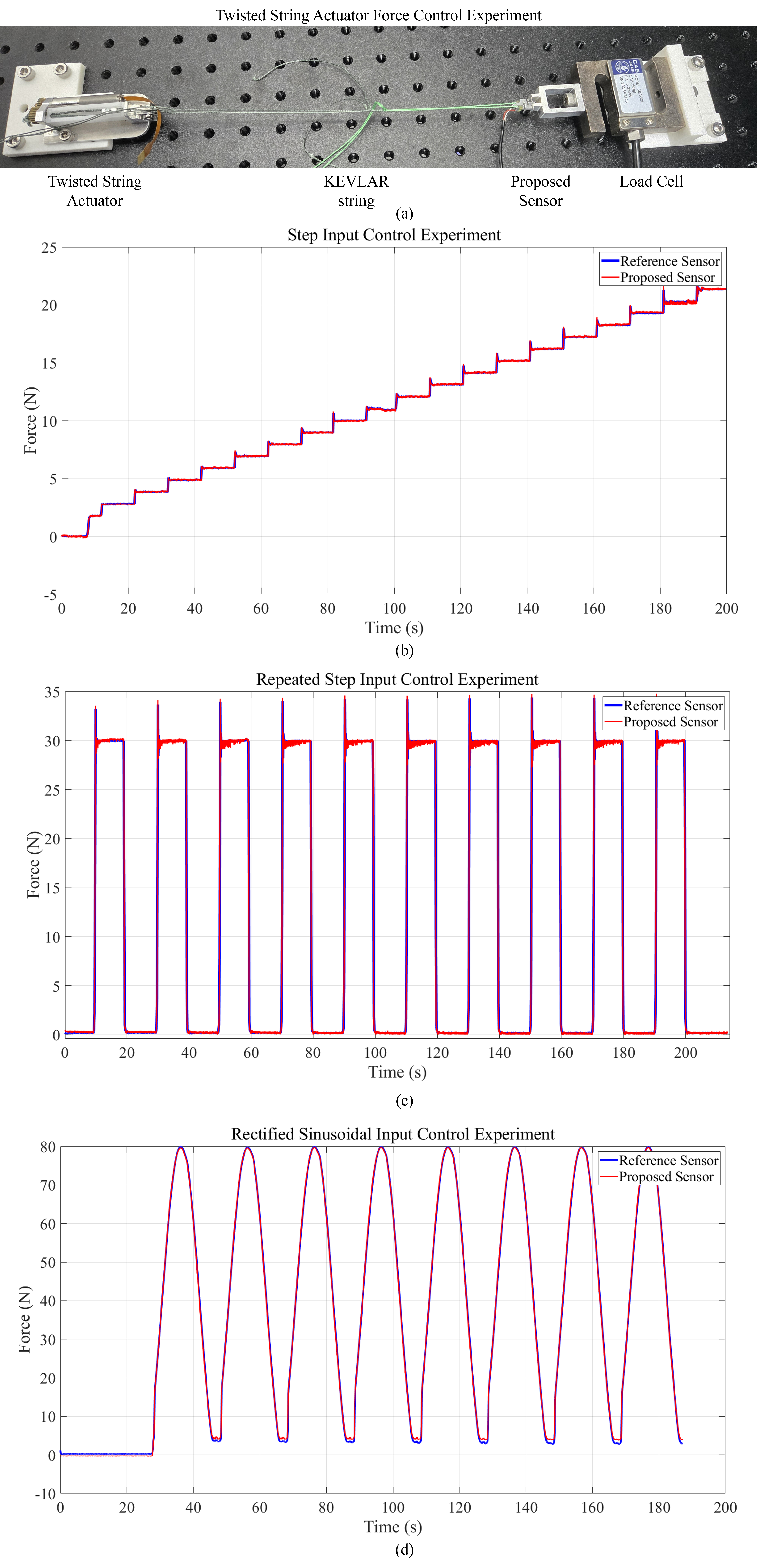}
    \caption{Control experiment setup and results: (a) Experimental setup for force control using a TSA, showing the motor, sensor, and control electronics; (b) Force control response to a step-wise increasing input, validating the system’s responsiveness and stability over multiple force levels; (c) Force control response to repeated 30~N step inputs, used to evaluate repeatability and control accuracy; (d) Force tracking performance under a rectified sinusoidal input profile, validating the sensor's capability in dynamic and continuous control tasks.}
    \label{controlexperiment}
\end{figure}
The calibration was performed by applying force using a Kevlar string. Since the target force for applications such as grippers or robotic hands is approximately 50~N, the calibration was conducted up to 70~N to ensure sufficient coverage. Due to the nonlinearity of the photo-reflector, a third-order polynomial was used, and calibration was performed using the least squares method. The results are shown in Fig.~\ref{calibrationexperiment}. Fig.~\ref{calibrationexperiment} (a) presents the calibration curve, while Fig.~\ref{calibrationexperiment} (b) shows a magnified view of the proposed sensor's output.

The results of the sensor calibration experiment are summarized in Table~\ref{resultproposed}. The RMSE was approximately 0.4550~N. Both nonlinearity and hysteresis, expressed as percentage errors relative to the maximum measurable force of 200~N, were approximately 0.80\% and 0.89\% as illustrated in Fig.~\ref{nonlinear} (a) and (b), remaining below 1\%. Furthermore, following the approach used in previous studies, the standard deviation of the sensor output over a stationary 10-second interval was calculated to be approximately 9.888~mN. This corresponds to a resolution of about 9.9~mN, which, when divided into the full scale of the sensor, yields 20,959 discrete steps-equivalent to a resolution exceeding 14 bits. Each quantization step of approximately 9.9 mN is visually confirmed in the enlarged region, as indicated by discrete plateaus in Fig.~\ref{calibrationexperiment} (b).
The sampling rate reached up to 5~kHz, consistent with previous research. As shown in Fig.~\ref{calibrationexperiment} (b), each step corresponds to approximately 9.9~mN. Notably, these results were obtained without applying any Kalman or low-pass filtering; thus, applying such filters is expected to further improve resolution.

\begin{table}[!htb]
    \centering
    \caption{Result of Proposed Sensor's Calibration Experiment}
    \begin{tabular}{cc}
    \hline\hline
    Performance Metric & Value \\ \hline
    RMSE &0.4450~N  \\
    Nonlinearity and Hysteresis     & 0.80~\%\\
    Standard Deviation & 9.888~mN\\
    Resolution & 9.888~mN\\
    Resolution (step) & 20,959 step(14$\sim$15~bit)\\
    Sample Rate & Max. 5~kHz\\
    \hline
    \end{tabular}
    
    \label{resultproposed}
\end{table}



Next, force control experiments were conducted using a TSA, which is commonly employed in robotic hands and grippers. The TSA is a high gear-ratio mechanism that utilizes strings to generate large forces with a small motor, and its inherent compliance makes it suitable for applications such as prosthetic hands. However, due to this compliance, TSA mechanisms are inherently difficult to control in terms of force using only position control.

As shown in Fig.~\ref{controlexperiment} (a), a Brushless DC(BLDC) motor from Maxon (EC series) was used in the experiment. Motor control was achieved by utilizing the BLDC motor output section of the setup introduced in Fig.~\ref{calexpins} (b). The motor driver used was the L6235 from STMicroelectronics, a compact driver capable of controlling BLDC motors via pulse width modulation (PWM). The motor operates at 12~V with an output power of 8~W.

Using this setup, force control was performed in three types of experiments, as illustrated in Fig.~\ref{controlexperiment} (b), (c), and (d). These include (b) step input tests with gradually increasing force levels, (c) repeated application of a 30~N step input over 10 trials, and (d) force control using a rectified sine wave function as the target input.
\begin{table}[!htb]
    \centering
    \caption{RMSE of Each Force Control Experiment}
    \begin{tabular}{cc}
        \hline\hline
        \textbf{Experiment Type}  & \textbf{RMSE (N)} \\\hline
        Step Input (a) & 0.0734\\
        Repeated Step (10 trials) (b) & 0.5953\\
        Rectified Sinusoidal (c) & 0.6103\\\hline
    \end{tabular}
    
    \label{resultexp}
\end{table}

The experiments were conducted using a PI controller. As shown in Table~\ref{resultexp}, Experiment (a) yielded a low RMSE of 0.0734~N. Experiments (b) and (c) showed RMSE values of 0.5953~N and 0.6103~N, respectively. These values are slightly higher than that observed in the calibration experiment, although Experiment (a) demonstrated even lower error.

\section{Discussion}
This paper presents a tension sensor utilizing a photo-reflector, designed for compact grippers and robotic hands. The proposed sensor, with a compact size of 13~mm, is capable of measuring forces up to approximately 200~N. A symmetric structure was adopted to reduce torsion compared to previous designs, and durability and sensitivity were enhanced by incorporating both a fillet structure and a flexure hinge. The elastomer was designed based on Timoshenko beam theory, and its behavior was verified through FEM analysis.

The PCB was fabricated using a photo-reflector and two resistors, and mechanically assembled using bolts. Calibration experiments were conducted, yielding an RMSE of approximately 0.445~N. The resolution, derived from the standard deviation, was approximately 9.9~mN, corresponding to 20,959 discrete steps-demonstrating a resolution exceeding 14 bits. This represents more than a tenfold improvement compared to the conventional resolution of 0.1~N. Such high performance was achieved through careful analog circuit design, including decoupling between analog ground and the 3.3~V analog signal using a large capacitor, and by performing measurements in the near region of the photo-reflector to reduce signal distortion. The system supports a sampling rate of up to 5~kHz, which depends on the resistance and capacitance associated with the photo-reflector.

Additionally, force control experiments were performed using a TSA with a PI controller under three test conditions. The corresponding RMSE values were 0.0734~N, 0.5953~N, and 0.6103~N, respectively.

\begin{table}[!htb]
    \centering
    \caption{Comparison between proposed sensor and Prior Work}
    \resizebox{0.5\textwidth}{!}{
    \begin{tabular}{ccc}
        \hline\hline
        Items & Prior Work & Proposed Sensor\\ \hline
       Maximum Allowable Force  &200~N &207~N  \\
Dimension&4~mm$\times$6~mm$\times$10~mm&6.5~mm$\times$7~mm$\times$13~mm\\
       Non-linearity/Hysteresis &1.85\%&0.8\%\\
       Resolution &0.1~N&0.0099~N\\
       RMSE  &1.36~N&0.45~N\\
       \hline
    \end{tabular}}
    
    \label{comparison}
\end{table}
Compared to conventional designs based on photo-interrupters, the proposed sensor achieves a tenfold improvement in resolution as shown in Table~\ref{comparison}~\cite{jeong2018design}. Furthermore, the symmetric elastomer structure contributes to reduced nonlinearity and hysteresis, demonstrating improved performance over previous methods.

In previous approaches, linear (first-order) fitting functions were used for calibration, whereas in this study, a third-order polynomial was applied. As a result, the previous method exhibited greater nonlinearity and higher RMSE. Although the proposed sensor has a slightly larger size due to the characteristics of the photo-reflector, it is still compact enough to fit within a fingertip. The maximum allowable force is also slightly higher in this study. Most notably, the resolution has significantly improved-from approximately 0.1~N in previous research to 0.009888~N in the proposed method-representing more than a tenfold enhancement.

However, as both photo-interrupter and photo-reflector systems rely on infrared light, they are inherently susceptible to temperature-induced errors. Prior studies have shown that infrared components exhibit a dark current characteristic, with voltage drift up to 0.35~V across a temperature range of $-20^{\circ}\mathrm{C}$ to $60^{\circ}\mathrm{C}$.~\cite{kim2025temperature} To mitigate this, future work will explore the use of temperature compensation using a gated recurrent unit (GRU)-based learning model.

Another limitation lies in the nonlinear voltage-distance relationship of photo-reflectors, which is more pronounced than that of photo-interrupters. Future research will involve more accurate modeling of this nonlinearity to further enhance sensor performance.
\section{Conclusion}
In summary, the proposed photo-reflector-based tension sensor offers high resolution, mechanical simplicity, and ease of integration, making it well suited for compact robotic systems. Its over-14-bit resolution and compatibility with standard analog acquisition systems highlight its practicality and scalability. The findings demonstrate that photo-reflectors offer a compact and manufacturable solution for achieving high-resolution tension sensing in robotic applications.

Future directions include integrating the proposed sensor into a robotic or prosthetic hand platform, enabling both position and force control using electromyography (EMG) signals~\cite{lee2024real}. Such an extension will support advanced wearable and assistive technologies, including soft robotics and human-robot interaction systems, where compact and accurate force sensing is critical.

\section*{Acknowledgment}
This work was supported by Agency for Defense Development(ADD) - Grant funded by Defense Acquisition Program Administration (DAPA) in UD230004TD.

\bibliographystyle{Bibliography/IEEEtranTIE}

\begin{IEEEbiography}[{\includegraphics[width=1in,height=1.25in,clip,keepaspectratio]{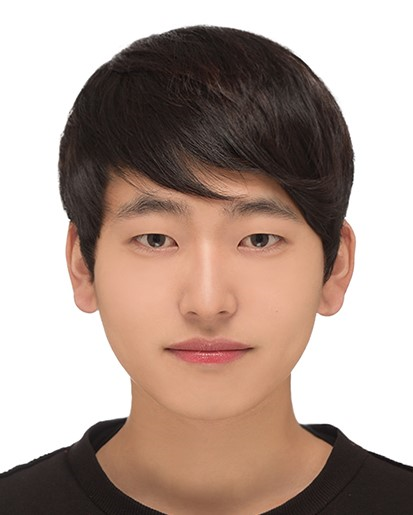}}]
{Hyun-Bin Kim}
~ obtained his B.S., M.S., and Ph.D. degrees in Mechanical Engineering from Korea Advanced Institute of Science and Technology(KAIST), Daejeon, Republic of Korea, in 2020, 2022, and 2025 respectively. Currently serving as a post-doctoral researcher at KAIST, his current research interests include force/torque sensors, legged robot control, robot design, and mechatronics systems.
\end{IEEEbiography}

\begin{IEEEbiography}[{\includegraphics[width=1in,height=1.25in,clip,keepaspectratio]{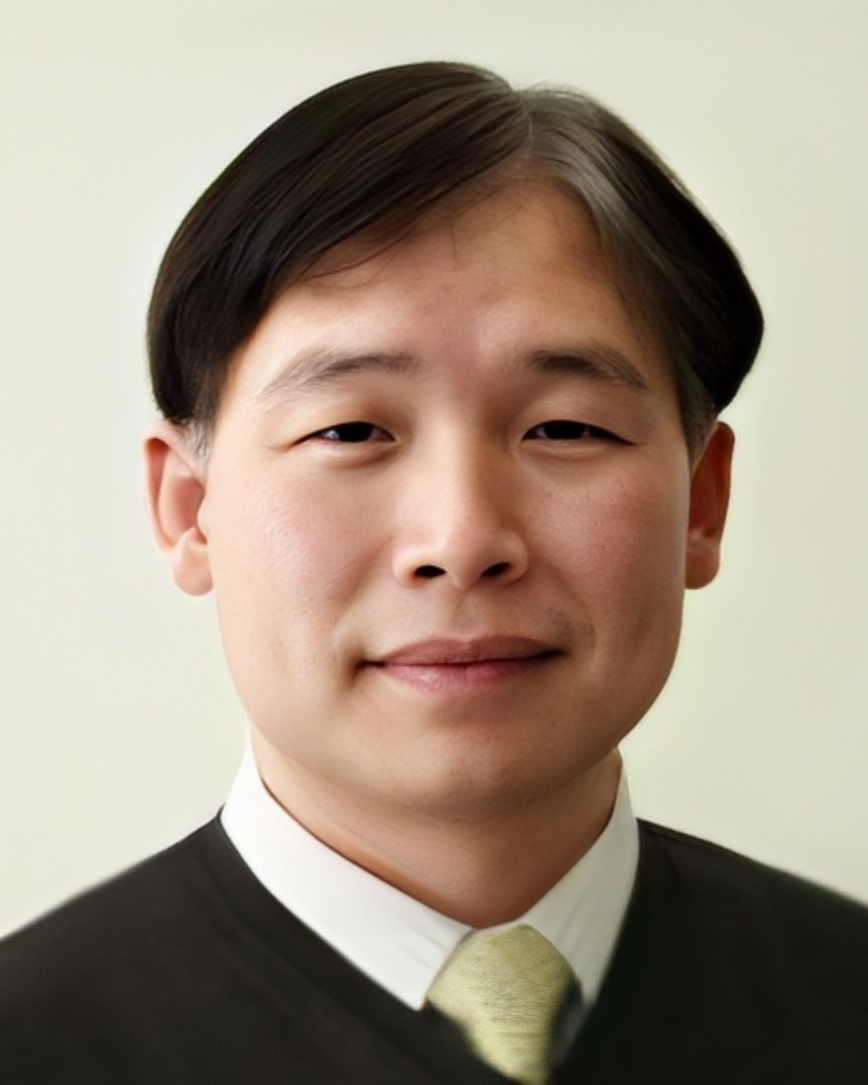}}]{Kyung-Soo Kim}~(Fellow, IEEE)~ obtained his B.S., M.S., and Ph.D. degrees in Mechanical Engineering from Korea Advanced Institute of Science and Technology (KAIST), Daejeon, Republic of Korea, in 1993, 1995, and 1999, respectively. Since 2007, he has been with the Department of Mechanical Engineering, KAIST. 
His research interests include control theory, electric vehicles, and autonomous vehicles. He serves as an Associate Editor for the Automatica and the Journal of Mechanical Science and Technology.
\end{IEEEbiography}

\end{document}